# Imaging and Classification Techniques for Seagrass Mapping and Monitoring: A Comprehensive Survey

**MD MONIRUZZAMAN,** Edith Cowan University, Australia
**S. M. SHAMSUL ISLAM,** Edith Cowan University, Australia
**PAUL LAVERY,** Edith Cowan University, Australia
**MOHAMMED BENNAMOUN,** The University of Western Australia, Australia
**C. PENG LAM,** Edith Cowan University, Australia

Monitoring underwater habitats is a vital part of observing the condition of the environment. The detection and mapping of underwater vegetation, especially seagrass has drawn the attention of the research community as early as the nineteen eighties. Initially, this monitoring relied on in situ observation by experts. Later, advances in remote-sensing technology, satellite-monitoring techniques and, digital photo- and video-based techniques opened a window to quicker, cheaper, and, potentially, more accurate seagrass-monitoring methods. So far, for seagrass detection and mapping, digital images from airborne cameras, spectral images from satellites, acoustic image data using underwater sonar technology, and digital underwater photo and video images have been used to map the seagrass meadows or monitor their condition. In this article, we have reviewed the recent approaches to seagrass detection and mapping to understand the gaps of the present approaches and determine further research scope to monitor the ocean health more easily. We have identified four classes of approach to seagrass mapping and assessment: still image-, video data-, acoustic image-, and spectral image data-based techniques. We have critically analysed the surveyed approaches and found the research gaps including the need for quick, cheap and effective imaging techniques robust to depth, turbidity, location and weather conditions, fully automated seagrass detectors that can work in real-time, accurate techniques for estimating the seagrass density, and the availability of high computation facilities for processing large scale data. For addressing these gaps, future research should focus on developing cheaper image and video data collection techniques, deep learning based automatic annotation and classification, and real-time percentage-cover calculation.

Authors' addresses: MD Moniruzzaman, Edith Cowan University, 270 Joondalup Dr, Joondalup, WA, 6027, Australia, mmoniruz@our.ecu.edu.au; S. M. Shamsul Islam, Edith Cowan University, 270 Joondalup Dr, Joondalup, WA, 6027, Australia, syed.islam@ecu.edu.au; Paul Lavery, Edith Cowan University, 270 Joondalup Dr, Joondalup, WA, 6027, Australia, p.lavery@ecu.edu.au; Mohammed Bennamoun, The University of Western Australia, 35 Stirling Highway, Perth, WA, 6009, Australia, mohammed.bennamoun@uwa.edu.au; C. Peng Lam, Edith Cowan University, 270 Joondalup Dr, Joondalup, WA, 6027, Australia, c.lam@ecu.edu.au.

## 1 INTRODUCTION

Seagrasses mostly live in marine sediments or rocky sub-tidal and inter-tidal habitats, where they often form meadows [16, 84]. Coastal areas around the world support upwards of 60 species of seagrasses which belong to four different families [102] (Figure 1). Seagrasses provide a number of important ecosystem services, including the provision of primary production supporting food webs, nutrient cycling, and habitat provision supporting marine biodiversity [36]. They are also a major sink of atmospheric carbon dioxide [29], helping to mitigate global climate.

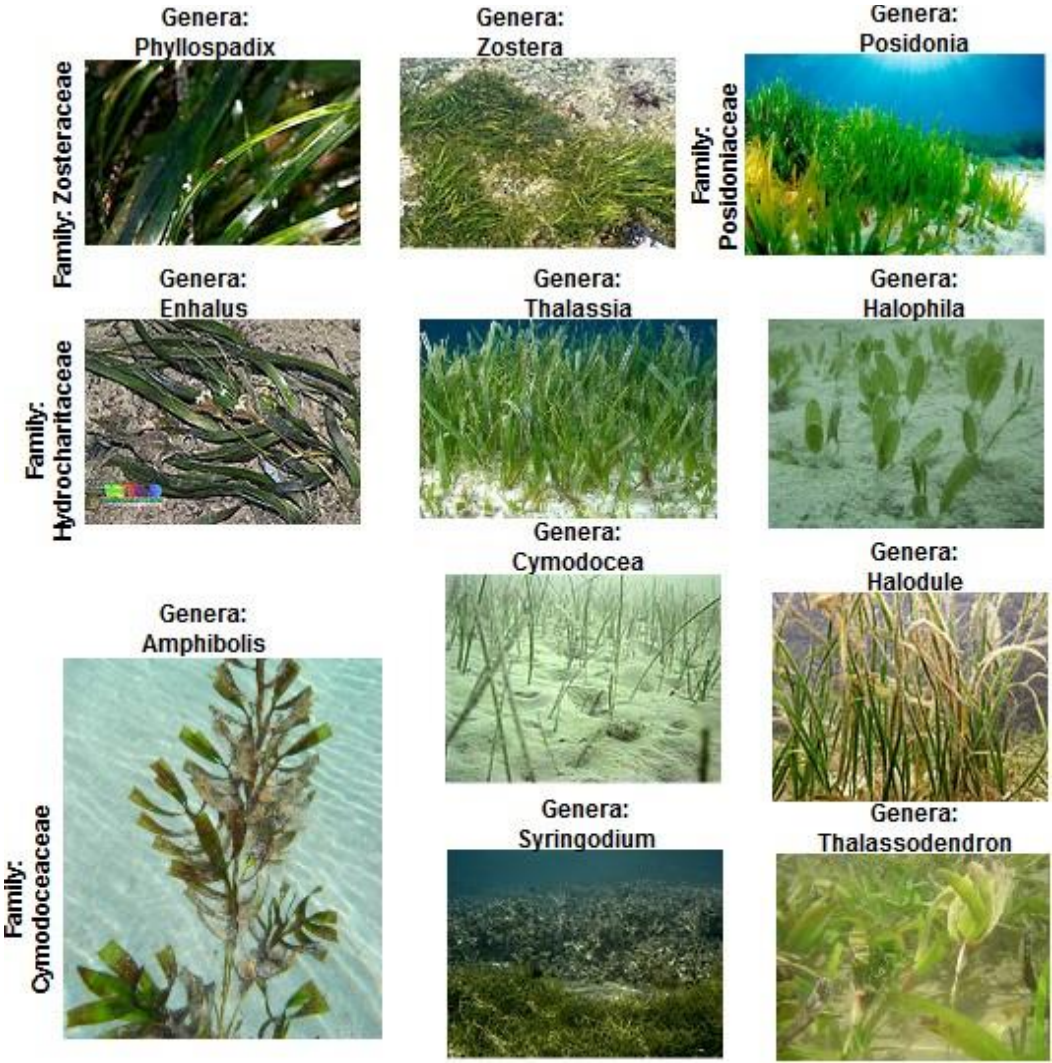

Fig. 1. Seagrass Types (Wikipedia, best seen in colour)

Despite their widely recognised importance, the rate of seagrass loss has increased significantly, mostly because of anthropogenic impacts but also from natural disturbance [97, 109]. Between 1980 and 2000, the rate of seagrass reduction increased to 110 square km per year, or in the order of 7% per annum [111]. Monitoring the extent and condition of seagrass is important not only to track the impact of human activities on this valuable resource, but because seagrass also acts as a sentinel, reflecting the condition of the surrounding environment, and so it is often used as a bio-indicator in monitoring programs [55, 57, 59].

The monitoring of seagrass is an arduous task. The prerequisite for monitoring is to detect and map seagrass distribution along the world's coasts. Maps with critical information about the extent, density, composition, and condition of the species can inform coastal management policies and plans, such as selecting MPAs (marine protected areas), prioritising actions addressing environmental decline [71].

Seagrass detection and mapping techniques vary, from in situ, diver-based surveys to the modern remote-sensing techniques. The initial methods and standards for studying seagrass were described by Walker [108] and Phillips & McRoy [83]. Later techniques and standards are recorded by English et al. [23], Coles et al. [13], Debson et al. [20], Kirkman et al. [52], Lee Long et al.[63] etc. All of these early documentations included either direct observations by divers or the extraction of coverage from aerial photography. The next stage in mapping extent took advantage of the flourishing remote-sensing technologies, high-resolution satellite imagers, lower-orbit and airborne earth-looking camera technology, which still strongly contribute to seagrass detection and mapping [71]. However, these technologies are largely applied to map or monitor the extent of seagrass. Many seagrass monitoring programs also require data on the cover of seagrass and the species composition at small spatial scales (<0.5 $m^2$) which even the best remote-sensing techniques cannot provide. The current advances in digital camera technology, autonomous vehicles, and high-performance computation facilities have widely opened the door for potentially easy and automated underwater vegetation detection and mapping methods.

Aside from data-collection methods, the media and types of the data collected, as well as their processing techniques and classification methods, can contribute to the automation of seagrass-monitoring procedures. The widely used image data types that are for detecting and monitoring seagrass include two-dimensional spatial and underwater images, underwater videos, acoustic sonar images, and spectral images mainly acquired from satellite remote sensing. Different image categories serve different purposes for monitoring seagrass. Close-view data types, such as underwater images and video, are useful for assesing the condition of seagrass, while spatial and spectral images mainly provide quick information regarding the presence and changes in the extent of seagrass.

Based on the image types, different detection and mapping techniques have been used starting from fully manual expertise-based approaches to supervised and unsupervised machine learning classification techniques. A few notable machine learning classification tools, such as support vector machine (SVM) [93], linear discriminant analysis (LDA) [115] and, principal component analysis (PCA) [18] etc. have been repeatedly used for seagrass classification. The advances in computational devices, data-acquisition techniques, and new classification tools provide scope to improve the detection, classification, and mapping of seagrass and advance the already established approaches to facilitate ocean monitoring.

In a previous survey, we addressed the application of deep neural networks as a classifier for marine object detections [74] in general, and seagrass in particular [43]. Kenny et al. [51] surveyed the seabed mapping techniques where their main focus was on sediment dynamics. Jana and Patrick [107] covered the machine learning based species identification approaches which included studies of fish detection and classification. However, there is currently no survey critically reviews seagrass detection, and classification approaches. Here we survey the advances in seagrass monitoring, discussing not only the data-collection methods but also the imaging approaches, data-processing, and classification techniques which can help researchers explore and understand the challenges and possibilities towards more autonomous and automated seabed mapping.

This article is organised as follows. Section 2 presents a taxonomy of state of the art seagrass detection approaches. Still image-based detection approaches are discussed in Section 3, video-based detection approaches are discussed in Section 4, acoustic-image approaches are covered in Section 5, spectral image data based approaches are covered and explained in Section 6. The challenges and gaps of the above approaches and future research needs are highlighted in Section 7. In Section 8 we present our conclusions considering the key constraints, performance, accuracy and promising approaches that have been discussed in the earlier sections.

## 2 TAXONOMY

We conducted a rigorous survey based on journal and conference articles, book chapters, review articles, theses, and project reports from online searching sources. We used databases such as Web of Science, Science Direct, ACM digital library, Google Scholar, IEEE Xplore digital library, Springer, Elsevier, and PubMed. Following keywords were used: 'seagrass', 'monitoring', 'detection', 'mapping', 'classification', 'image processing', 'machine learning', 'underwater image', 'satellite image', 'aerial photography', 'acoustic image', '3-D seabed', 'spectral image', and 'underwater vegetation'.

Based on the literature review, we categorise seagrass detection and monitoring approaches from two perspectives, either by asking what type of data were collected or how the data were collected (e.g., satellite, aerial photography). Initially, all the methods are divided based on four different data types: still image-based; video footage-based; acoustic data-based; and spectral image-based approaches (Fig. 2). The still image based approaches are further sub-divided into spatial image based approaches (section 3.1) and underwater image based approaches (section 3.2) as these two types of image data possess their own distinctions in terms of data processing, and applications. Approaches using spatial still photographs are further sub-divided into two categories: analogue image based approaches (Section 3.1.1) and digital image based approaches (Section 3.1.2). The underwater image-based approaches are further sub-divided as 2D (Section 3.2.1) or 3D (Section 3.2.2) image-based approaches.

Based on the imager used for data collection, spectral image based approaches can be categorised into: aerial imager based approaches (Section 6.1), satellite imager based approaches (Section 6.2), portable imager based approaches (Section 6.3) and those approaches which use multiple types of imagers for data collection (Section 6.4). Satellite -acquired imagery can further be divided into: Landsat image based approaches (Section 6.2.1), IKONOS image based approaches (Section 6.2.2), QuickBird image based approaches (Section 6.2.3), and WorldView image based approaches (Section 6.2.4).

## 3 STILL IMAGE-BASED APPROCHES

### 3.1 Spatial Image-Based Approaches

Among the seagrass mapping methods of the last few decades, two-dimensional spatial or aerial photography-based visual interpretation remains the most adopted approach worldwide [103]. Aerial photographs are taken from specially equipped aircrafts, which maintain steady altitudes while flying. Normally, cameras are mounted under the fuselage and can be adjusted according to the requirements during the photo runs. Each aerial photograph overlaps with the immediate neighboring photo by 60% to facilitate the interpretation of the most central part of the photo, allow stereoscopic interpretation, and moreover compensate the loss of the coverage from sunglint. To ensure contiguous coverage, a 30% side lap is also maintained [71]. All the approaches using spatial images can be divided into two categories: analogue image based and digital image based approaches.

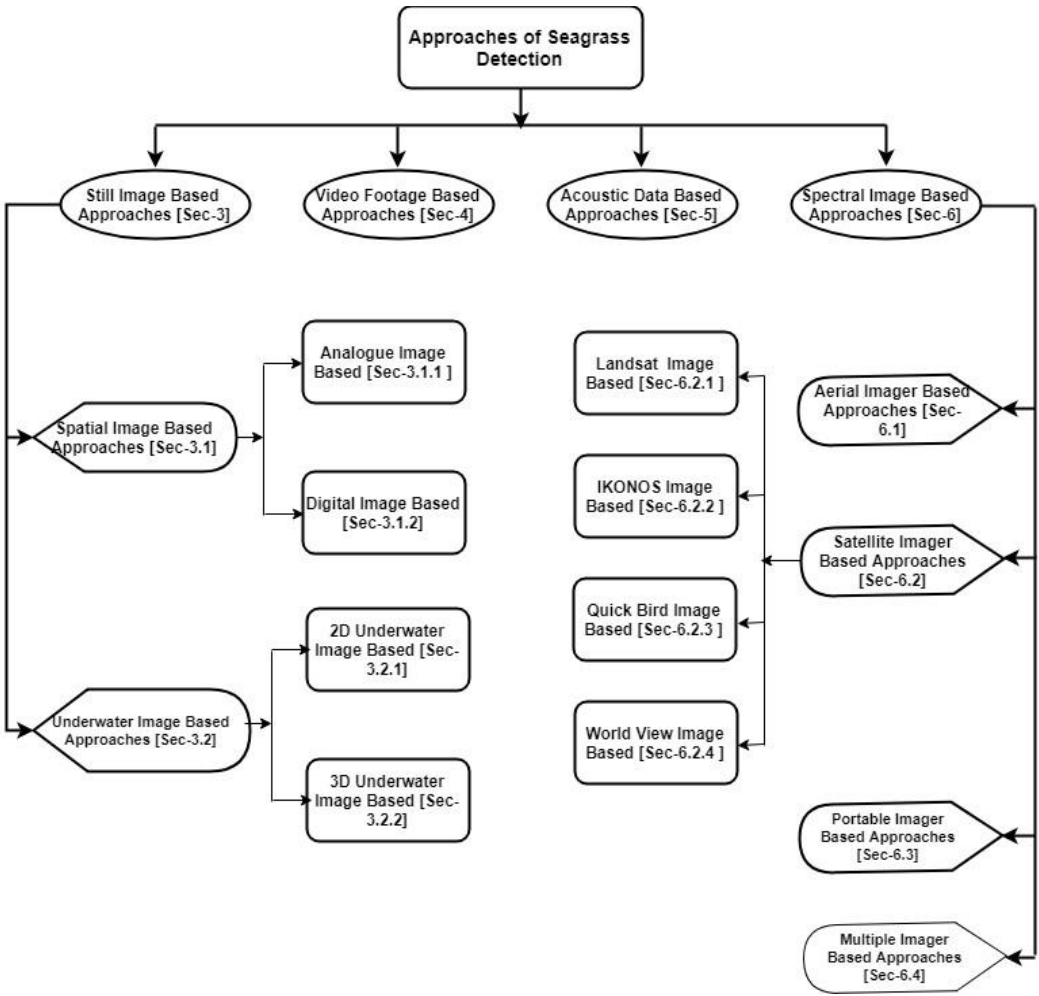

Fig. 2. Approaches of Seagrass Detection

*3.1.1 Analogue 2D Spatial Image-Based Approaches.* Some of the earliest and prominent approaches of using analogue aerial photographs to detect and map seagrass were performed by Carraway and Priddy (1983)[9], and Ferguson et al. (1993)[26]. In their approach of creating an accurate and detailed map of seagrass beds in the Bogue and Core Sounds in Carteret County, North Carolina, Carraway and Priddy [9] considered using the conventional, analog, two-dimensional natural-colour photographs taken by an aircraft. Manual survey-based ground truthing was performed for this approach. The photographs were printed on a nine-by-nine-inch paper format scaled one thousand feet per inch. Finally, a map of a modified scale (two thousand feet per inch) was made from accumulating all the photographs. Image interpretation, ground truth data addition, density confirmation were done manually.

To map the spatial change of seagrass in the black and southern Core Sound of North Carolina, Ferguson et al. (1993) [26] used two-dimensional analog photographs taken over the years between 1985 and 1988. They have added and accumulated the habitat information from the maps by the National Oceanic and Atmospheric Administration to their photographs. For error rate reduction, they emphasized on surface level training and added restrains to the photography for optimal visualisation of the images. This approach was limited to measuring changes of the seagrass bed extent only.

*3.1.2 Digital 2D Spatial Image-Based Approaches.* Digital camera technology and images have brought a new era of photography based seagrass detection, classification, and mapping techniques. One of the earliest of these approaches was performed by Chauvaud et al., [11]. While mapping seagrass beds, corals, and mangroves in the Bay of Robert at the French West Indies (Martinique Island), Chauvaud et al., [11] relied on two-dimensional spatial images taken from an aerial platform. They digitised those true-color photographs and separated the images into red, blue, and green bands. Afterwards, an unsupervised classifier (PCA) was applied. Based on the range of depths, various masks were built and applied to the original image. This new image was then classified using a supervised classification (maximum probability algorithm) technique. For each image, both classification techniques were merged, and a final map was produced.

Kendrick et al. [49, 50] used both 2D aerial images of three colour bands and video footage from a towed camera by boat to create a coverage change map for the period of 1967 to 1999 in Cockburn Sound, and Success bank, Western Australia. Image processing was performed by Geographic Resources Analysis Support System (GRASS) and a map was produced using a semi-automated greyscale segmentation method named Spann-Wilson [98] method. Manual video based survey [48] was performed for validation and ground truthing. Similarly combination of 2D spatial image based mapping with video based validation was performed by Holmes et al. [38].

Paulk [82] described a technique where they used a feature-classification algorithm to detect and map *Posidonia oceanica* east of Pythagorio, Greece using unspecified satellite images. The authors used ENVI software and the nearest neighbour classification algorithm incorporated into this software to generate polygons containing *P. oceanica*. To create a complete map, Paulk used ArcGIS software. The classified polygons were feed as layers, and then depth contours were added. This classification and mapping procedure is mostly a semi-automatic approach which requires two distinct types of software. The accuracy calculation totally depends on ground truthing by an in-situ survey. To construct a complete map of a region, the main challenge is the collection and integration of images which provide a set of data points to complete the map. Though satellite images have been used for this approach, it has been included under the category of two-dimensional spatial image-based category, as the photographs used for classification are aerial images that had visual seagrass (*Posidonia oceanica*) data points and were super imposed upon the satellite images of the experiment site.

The later improvement of seagrass mapping from digital image data used a linear spectral unmixing of aerial imagery dataset. This approach was experimented by Uhrin and Townsend [103]. They examined whether a Linear Spectral Unmixing classifier (LSU) could improve the detection accuracy, compared with the manually digitised seagrass beds using established protocols. The optimal pixel proportion was assessed by the Euclidean distance from ROC curve (receiver operating characteristics). The area under the ROC curve (AUC) and kappa statistics parameters were used to measure the performance of the accuracy. After the experiment, they found that LSU performs better than the manual digitisation method. Table 1 lists all the two-dimensional spatial image-based approaches discussed above.

## 3.2 Digital Underwater Image based Approaches

As the underwater camera technologies improve along with cheaper storage and the availability of high-computation facilities, underwater digital still images can be the preferred image data source for underwater vegetation mapping [74]. Although very few approaches were performed using underwater image datasets, both 2D and 3D image types have been utilised based on the specifications of the mapping. Both 2D and 3D underwater image based approaches are discussed separately below.

Table 1. List of seagrass detection approaches based on 2D spatial images

| Author | Data Type | Location | Source | Classifier | Accuracy |
| --- | --- | --- | --- | --- | --- |
| Uhrin & Townsend [103] | 2D Spatial Image | Albemarle-Pamlico Sound Estuary | Aerial Photo | Linear Spectral Unmixing (LSU) | 86.3-99.0% |
| Paulk [82] | 2D Aerial Imagery | Region east of Pythagorio | Satellite Image | Nearest Neighbour Feature Classification Algorithm | N/A |
| Kendrick et al. [50] | 2D Spatial and Video | Cockburn Sound, WA | Aerial photo & towed camera | Spann-Wilson segmentation | NA |
| Chauvaud et al. [11] | 2D Spatial Image | Bay of Robert, French West Indies | Aerial Photo | PCA & Maximum Probability Algorithm | 86-94% |
| Ferguson et al. [26] | 2D Spatial Image | Black and southern core sound of North Carolina | Aerial Photo | Optimal Visualization technique | N/A |
| Carraway and Priddy [9] | 2D Spatial Image | Bogue and Core sounds in Carteret County | Aerial Photo | Manual Expertise | N/A |

*3.2.1 2D Underwater Image Based Approaches.* One of the earliest 2D underwater digital image based approaches for seagrass detection was performed by Yamamuro et al. [116]. Using a remotely operated vehicle (ROV), they developed a mapping system from seabed photos (Fig. 3). Those photographs were filtered through the green light channel, and the coverage calculation was performed using semi-automatic digital image processing based on the Otsu classification method. To gather accurate results from the ROV-based dataset, image locations must be incorporated with the digital image data. Moreover, ROVs are an expensive mean to collect underwater digital data due to their initial capital cost, although modern technology has reduced the operational cost for ROVs significantly.

Pizarro et al. [87] experimented with a technique that is very similar to the 'bag-of-words' approach. They used a bag-of-features for object recognition from 2-D underwater images and applied it to three different scenarios: digital still photos from towed camera; fast summery-function application for AUV surveys; and adaptive habitat mapping via AUV. Bag-of-features techniques considers images as a collection of independent patches. A visual descriptor vector is calculated for each patch, and each image is described by the distribution of the samples in the descriptor space [80]. While working with towed-camera images, they trained and tested their machine-learning algorithm with 453 expert annotated images, which were divided into eight classes, each class containing 52 to 63 images. They applied 10 iterations for this approach. Their algorithm showed a noticeable and significant confusion between some classes. They carried out an experiment at Jervis Bay, Australia, using an AUV for adapting habitat mapping using the bag-of-features algorithm. They trained their algorithm with 20 photographs of eight different classes and implemented it in the AUV, which created an initial habitat map. Afterwards, the classifier was applied to validate the map created initially by the AUV. For validation, Pizarro et al. [87] used 1,860 images collected by AUV during the survey and finally created an interpolated map. Accuracy for the classification task of this approach has not been considered while validating the map.

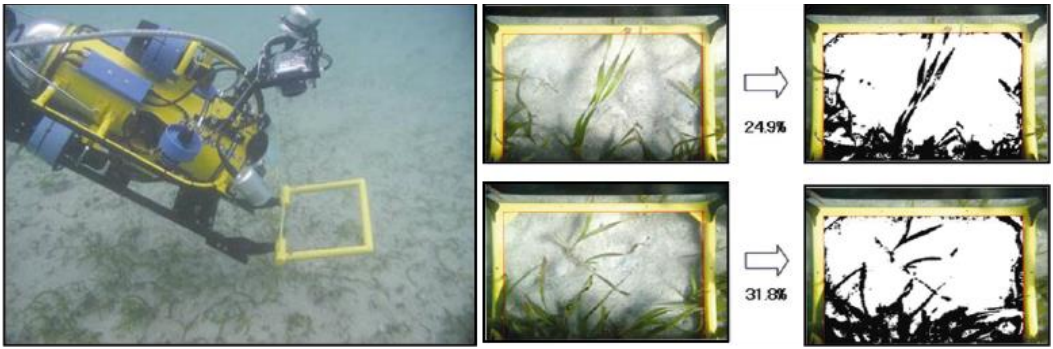

Fig. 3. Seabed digital photo collection using ROV and semi-automated calculation of seagrass coverage (%) (Yamamura et al., [116]).

Using underwater two-dimensional colour photographs of benthic habitats, Jalali et al. [44] tested several machine-learning classification models, such as support vector machine (SVM) classifier based model, hierarchical max (HMAX) model, and colour-quantization hierarchical max (CQ-HMAX) model. Though this approach did not directly aim to detect and classify seagrass from underwater images, it detected Sargassum and other seaweeds from a dataset which contains 19 different classes of benthic oceanic organisms, each having 60 to 300 two-dimensional colour photographs. The authors initially resized the photographs to 500 × 300 pixels without changing the aspect ratio. While applying SVM, the dataset was further resized to 160 × 90 pixels, and for HMAX and CQ-HMAX, the conversion was 140 × Si (Si relies on the aspect ratio). In the first layer of the hierarchical architecture, the HMAX classifier used Gabor filters, feature extraction, and pool features and improved the classification accuracy (48.22±1.8%) with respect to SVM (20.54±1.8%). Instead of the Gabor filter, CQ-HMAX used colour quantisation which increased the accuracy to 56.23±0.5%. As a final approach, Jalali et al. [44] combined HMAX and CQ-HMAX to use both shape and colour features (Fig. 4), and that increased the accuracy to more than 61%. Though this approach gave a positive indication of the usability of machine-learning classifiers to detect seagrasses from 2-D images, it failed to distinguish seagrass from Sargassum and other seaweeds by classifying all as seagrass. Additionally, the rate of accuracy was also measured in contrast to other colourful benthic habitats such as corals, sponges, and tubulates. Therefore, the classification of different species of seagrass would impose a further challenge for this classifier, since the colour difference is almost negligible between seagrass classes.

Massot-Campos et al. [68] used digital underwater images and applied multiple machine-learning algorithms for their seagrass-detection technique. They analysed the texture of the seabed digital images to detect and quantify *Posidonia oceanica* at Palma Bay, Spain. Three thousand RGB 720×576 pixel images were taken from an underwater platform. Every image was divided into several non-overlapping patches which were used to train the classifier (Fig. 5). For texture-difference analysis, they used the grey-level co-occurrence matrix (GLCM) approach and Law's energy measurements. They used machine-learning algorithms on this texture data to quantify *Posidonia oceanica*. They experimented with the multilayer perceptron (MLP), random forest tree (RF), and logistic model tree (LMT) classifiers and finally chose LMT for its good classification rate and simplicity. As it is only a texture-based machine-learning approach, the detection may easily provide erroneous results in seagrass detection if the image contains algae or seaweed because of their almost similar textures.

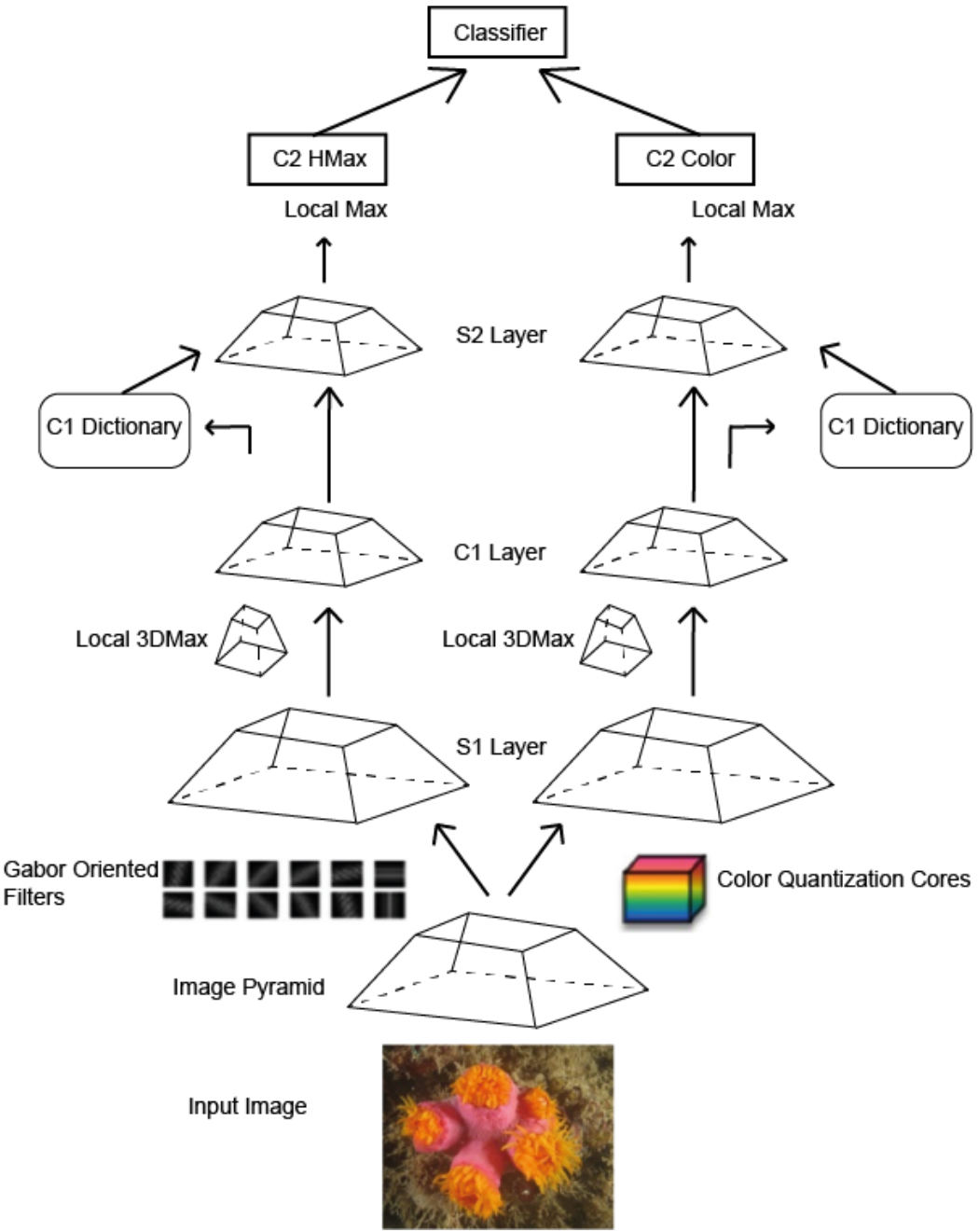

Fig. 4. Integration of CQ-HMAX with HMAX model (Jalali et al. 2013) [44]).

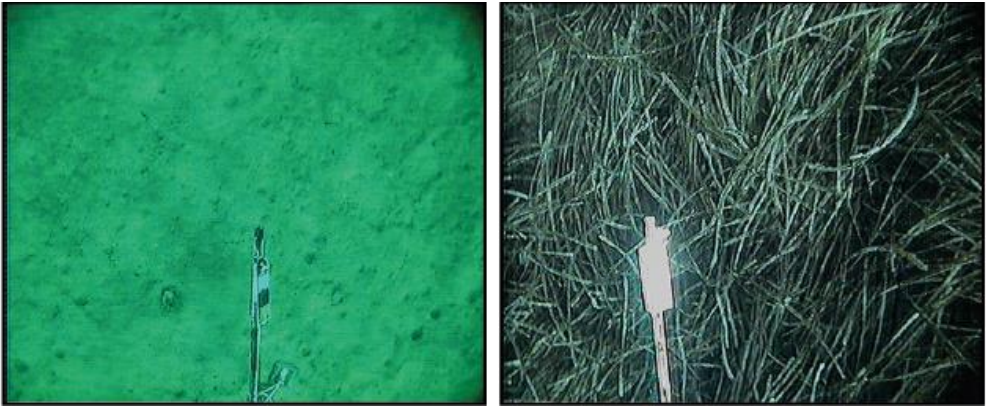

Fig. 5. An underwater digital image of a seabed (no PO) and all PO define and their texture differences (Massot-Campos et al. [68]).

Towards a more autonomous and automatic seagrass detection and classification process, Burguera et al. [7] applied different image-processing techniques to a dataset of 69 underwater still images collected by an AUV from the coastal areas of Mallorca. While processing the image dataset for classification, experimentation was done on local colour correction (LCC) [75], multiscale retinex (MSR) [45] and multiscale retinex- new kernel (MRS-NK) [76], MAI method [7], and Tone mapping algorithm [56]. For final image classification and *Posidonia oceanica* detection, the authors used a support vector machine (SVM) classifier. The classification accuracy was 92.9% for this approach. The highest classification was obtained on images pre-processed by the MAI method. This approach mainly focused on the image pre-processing techniques, while the main classifier was the same. Changing classifiers along with the image-quality enhancer would complement the whole approach.

Another approach for the identification of *Posidonia oceanica* from underwater digital images using machine-learning and deep-learning algorithms was proposed by Gonzalez-Cid et al. [32]. The images were collected using bottom-looking underwater cameras. In this study, the authors used two datasets. Dataset 1 consists of 69 colour images, and dataset 2 has 180 images. In their machine-learning approach, they used SVM and artificial neural network (ANN) algorithms. The role of SVM was to find the best hyperplane which differentiates all the data points from '*Posidonia oceanica* present' to '*Posidonia oeanica* not present'. During this machine-learning approach, the original images were converted to 640 × 480 pixels. Further, the images were subdivided into a set of 400 sub-images of 32 × 32 pixels. To train the SVM and ANN, texture descriptors and co-occurrence matrices were used. They also used GLCM after rescaling the images into eight gray-levels. For the deep-learning approach, they used an architecture inspired by the architecture used for CIFAR-10 dataset classification. To train the CNN architecture, they divided their images into patches of 32 × 32 pixels. All the datasets were divided into an 80% training set and a 20% test set, and each epoch was set to five hundred iterations. From this comparative approach, the authors found that the deep-learning algorithm increased the detection hit ratio by 1.54%. However, this approach is significantly slower than the SVM based machine-learning approach. The dataset size is very small in this experiment for a conventional deep-learning algorithm. Also, the number of iterations was set too high which could make the training approach saturated and time-consuming. Moreover, this approach does not classify different seagrass types, and instead gives a result from *Posidonia* and non-*Posidonia* images. Table 2 is a compilation of all the seagrass-detection approaches that are based on two-dimensional underwater images.

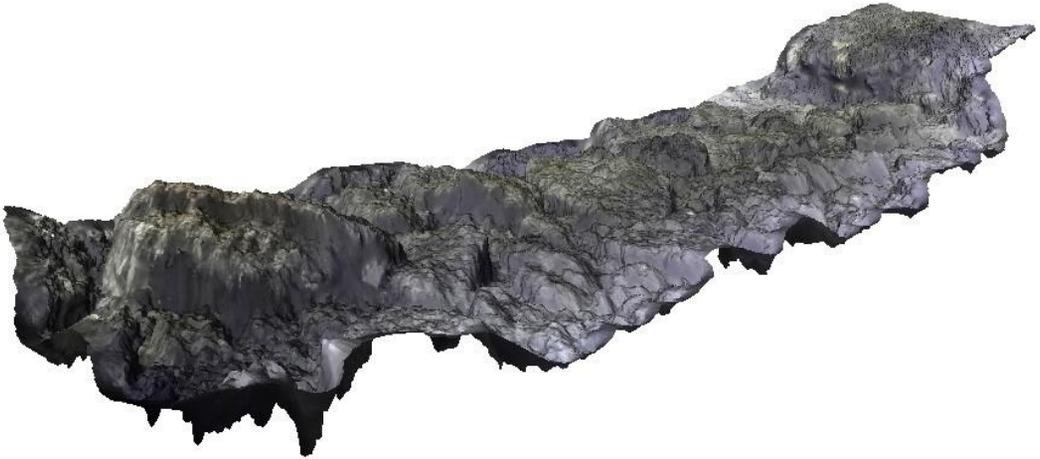

Fig. 6. Final constructed 3D model of a seabed transect (Rende et al. [90]) (Best seen in colour)

*3.2.2 3D Underwater Image Based Approaches.* To the best of our knowledge, the only approach in the underwater digital image category that has used 3D images of the seabed to detect seagrass (*Posidonia oceanica*) was performed by Rende et al. [90]. A pilot program to evaluate the performance of 3D model-based seagrass detection was undertaken at Capo Rizzuto, Italy. For image collection, a towed-camera system was used with a GoPro Hero 3+ 3D camera system mounted in a downwards posture. In this survey, the camera was set to a 108-degree FOV (field of view) and 1,080-pixel image quality. An appropriate positioning of the image was ensured using a GPS data logger. Deblurring, colour correction, and histogram stretching were performed to obtain an enhanced image. Afterwards, a dense stereo algorithm (multi-view) [30] was used to construct the 3D model of the seagrass meadow (Fig. 6). Using all the reconstructed 3D models, a point cloud was produced using an open-source software called Meshlab. Finally, the texture was added to the model. This model is a successful implementation of 3D image data to construct a seabed map. However this approach lacked the final part of seagrass detection or mapping as it did not extend to the automatic classification of seagrass from the 3D construction of seabed. Table 2 lists all the above approaches.

## 4 VIDEO-BASED APPROACH

Though few approaches used video-graphy to detect and map seagrasses, this modality has the potential to be useful for detecting and estimating cover, especially for deep-water seagrass meadows [79]. Video data-based techniques have proved credible for benthic distribution modelling (Holmes et al. [39]), marine animal abundance estimation (Auster et al. [5]) and benthic habitat cover calculation (Foster et al. [28]; Whorff and Griffing [113]).

Table 2. List of seagrass detection approaches based on underwater images.

| Author | Data Type | Location | Source | Classifier | Accuracy |
|---|---|---|---|---|---|
| Gonzalez-Cid et al. [32] | 2D Image (Underwater) | N/A | Bottom Looking Underwater Camera | SVM & ANN | ANN hit 1.54% higher than SVM |
| Burguera et al. [7] | 2D Still Imagery | coastal areas of Mallorca | AUV | SVM | 92.9% |
| Jalali et al. [44] | 2D Underwater Image | N/A | Unerwater Camera | SIFT, SVM, HMAX, CQ-HMAX | SVM (20.54 ±1.8 %), HMAX (48.22 ±1.8 %), CQ-HMAX (56.23 ±0.5 %) |
| Massot-Campos et al. [68] | 2D Underwater Image | Palma bay, Spain | Underwater platform | Logistic Model Tree (LMT), Random Forest tree (RF) and Multilayer Perceptron (MP) | N/A |
| Pizarro et al. [87] | 2D Underwater Image | Jervis Bay, Australia | AUV and Towed Camera | Bag of features based maximum likelihood classifier | N/A |
| Yamamuro et al. [116] | 2D Underwater Image | N/A | Remotely operated vehicle (ROV) | Otsu classifier | N/A |
| Rende et al. [90] | 3D underwater image | Capo Rizzuto, Italy | GoPro Hero 3+ 3D camera | Human expertise from 3D model | N/A |

One of the first applications of video footage for seagrass mapping was performed by Norris et al. [79] to quantify sub-tidal seagrass cover at Picnic Cove, Shaw Island, Washington (USA). Norris et al. [79] combined a geographic information system (GIS) and differential global-positioning system (DGPS) with a video graphic data-collection process. For video footage capture, a SeaCam 2000 was used and mounted downwards in a towfish deployed from the survey vessel. A video footage of the survey area was stored on a four-headed VCR (Video Cassette Recorder). Survey notes and back-up data were also stored in floppy disks at two-second intervals. The video footage was then usually interpreted by human experts for the presence or absence of seagrass and all the information was logged into a spreadsheet with other survey pieces of information such as DGPS and GIS positioning. The density of basal seagrass meadows was classified as low, medium, or high by human experts. Finally, the authors entered that spreadsheet information into a CAD program to produce a thematic map [79]. This approach is more suitable for the basal area coverage mapping of seagrass where aerial still photography and acoustic imaging fail to survey accurately. Moreover, the incorporation of DGPS and GIS data provide a higher confidence in positioning accuracy. However, the post-processing and density coverage calculation relies on human interpretation of video transects footage, which can be time-consuming and prone to observer errors.

Though not a direct use of video data for seagrass mapping, Mishra et al. [73] used video data to validate their benthic habitat mapping approach via IKONOS satellite images at Roatan Island, Honduras. For their model validation, they used a towfish technique where a towed sensor platform was used (Fig. 7). This is similar of deploying a video camera by Norris et al. [79]. The towfish was designed to resist wave action and move almost horizontally when towed at 3 km/h speed. For in-situ video capturing, they used a Sony Hi-8 mm TRV-320 digital camera. This approach did not apply any post-processing or classification technique of the video data nor was these any attempt made to calculate seagrass coverage from the video data.

A combination of video and sonar acoustic image based technique was performed by Lefebvre et al. [58]. In this approach, the use of video footage to detect and classify benthic habitats (*Zostera marina*, macroalgae and seabed) was limited to the validation of the acoustic experiment. The video footages were collected by a bottom facing high definition colour CCD camera mounted on an aluminium sledge. The sledge was towed by a boat at a speed of 1-2 knots. Footages were saved and played back for the manual detection and classification by experts. From replying the videos, the authors classify the vegetations and estimated the *Zostera marina* density into no coverage, sparse coverage, patchy coverage, dense coverage, and continuous coverage. This approach used a manual data interpretation technique for video data analysis and used this interpretation to validate the survey performed by sonar equipments.

Another approach which used a towed underwater video camera was proposed by Stevens et al. [100]. They recorded underwater seabed videos on a SONY Digital 8 'Handycam' mounted inside a PVC underwater housing hung from a survey vessel in Moreton Bay, Australia. From 78 sites of the bay, the authors collected 40 kilometers of video footage containing 16,373 individual frames. From each video frame, they counted all discrete and single seagrass organisms and calculated the percentage cover using a nine-point array. Extracting the counting data from the transacts, they implemented the Bray-Curtis dissimilarity and constructed similarity matrices. In this approach, the authors focused on manual counting-dependent coverage calculation and a human expertise-dependent classification technique. This approach is very similar to the approach that was performed by Norris et al. [79] and differs only in the equipment used for the video data collection.

Lirman and Deangelo used a shallow-water positioning system (SWaPS), whereby a video camera is equipped with a GPS receiver. This SWaPS was developed by the National Geodetic Survey of the NOAA (National Oceanic and Atmospheric Administration). This video-monitoring system was available in a diver-based platform (Fig. 8 (a) and (b)), a remotely operated platform, and a boat-based platform. They surveyed Black Point, Biscayne Bay, Florida, in depths less than one metre and at distances up to five hundred metres from the shoreline. While surveying, they used a spatial grid where each locational point was 150 metres apart. Each video segment captured a twenty-five-metre transect. From the video data of each transect, Lirman and Deangelo [61] chose ten random, non-overlapping frames (Fig. 8 (c)). From these frames, they determined the species of seagrass. For the percentage-cover calculation, they averaged the benthic coverage data from those frames and developed surface contours using the inverse distance-weighted interpolation method of the ArcView software. All the systems collecting seagrass images uses GPS to locate the images, they all require above water antena which is a significant limitation. Table 3 lists all the video footage based seagrass mapping approaches that we covered in this article.

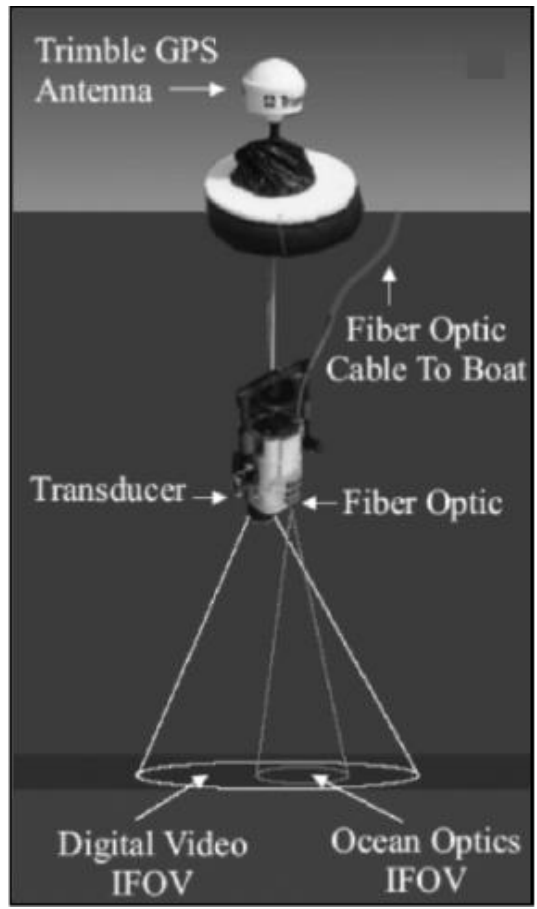

Fig. 7. Towfish equipped with a video camera and a GPS system (Mishra et al. [73]).

Table 3. List of seagrass detection approaches based on underwater video data

| Authors | Data Type | Location | Source | Classifier | Accuracy |
|---|---|---|---|---|---|
| Noriss et al. [79] | Video | Picnic Cove, Shaw Island, Washington (USA) | SeaCam 2000 | Manual | N/A |
| Mishra et al. [73] | Video | Roatan Island, Honduras | Sony Hi-8mm TRV-320 digital camera | Manual | N/A |
| Stevens et al. [100] | Video footage | Moreton Bay, Australia | Towed Handycam | N/A | N/A |
| Lirman and Deangelo [61] | Video | Black Point, Biscayne Bay, Florida | SWaPS | ArcView software | N/A |

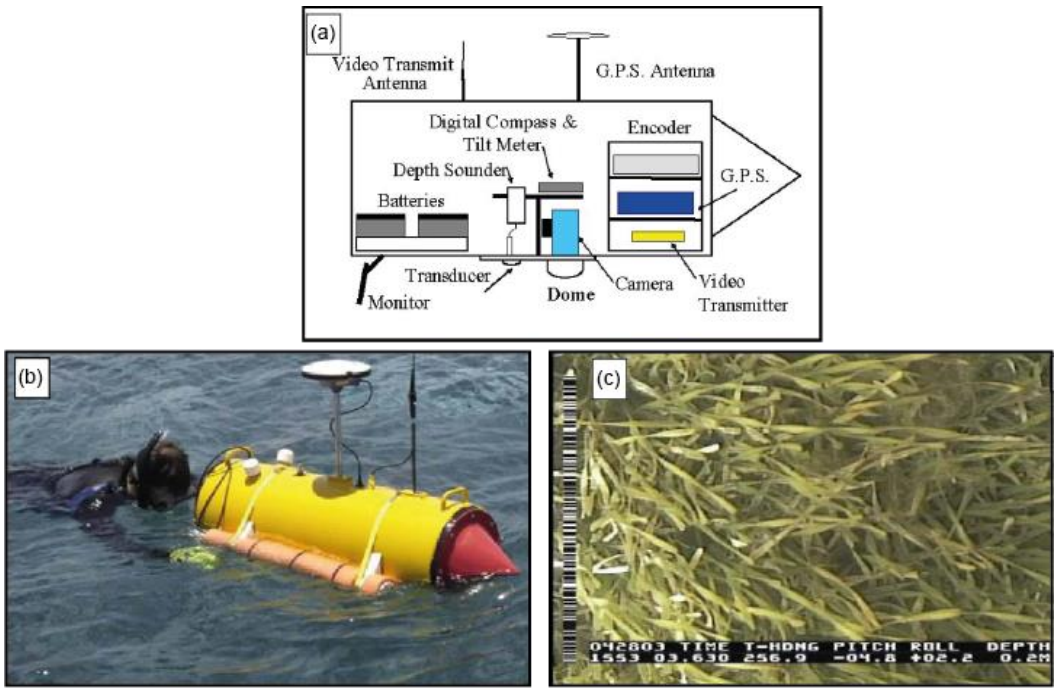

Fig. 8. (a) Diver-operated SWaPS platform schematic. (b) Video survey at Biscayne Bay, Florida. (c) Video frame of seagrass mapping using SWaPS (Lirman and Deangelo [61]).

## 5  ACOUSTIC IMAGE BASED APPROACHES

The number of attempted approaches is a clear indication that satellite spectral image or aerial still image surveys are the preferred methods of the research community, for their wide-spatial coverage capability. Turbidity, cloud coverage, algal bloom, tidal surge, or image distortions can create limitations and limit the outcomes for spectral images [106]. Video images are very promising solution to some of the issues discussed in Section 4, but sonar or acoustic image-based approaches overcome some of the limitations of video imagery approaches. Multi-beam sonar, side-scan sonar, or echosounder can be used to detect seagrass [114]. When using a sonar, the contrast of the density between an object and seawater creates a noticeable backscatter to the acoustic energy [47]. Because seagrass species, have air-filled tissues [94], the scattered sound energy or echo is stronger and can be used to detect and locate seagrass meadows [110]. A number of studies have used acoustic signals to detect seagrass instead of acoustic images such as, Maceina and Shireman [65], Duarte [21], Spratt [99], Miner [72], Hundley and Denning [40], and Sabol et al. [92]. Descamp et al. [17] used acoustic telemetry for their approach and a combination of side scan sonar acoustic data and Infrared spectral imagery was used by Pizzai et al. [86].

Komatsu et al. [53] used multi-beam sonar to detect and measure the volume of *Zostera caulescens*. In Otsuchi Bay, Sanriku, Japan, multi-beam sonar was used to create three-dimensional acoustic images of the seagrass meadows and a hydrography software to estimate the area and volume of *Zostera caulescens*. The validation of the experiment was performed by quadrant survey samples.

In their approach, Jones et al. [46] used a REMUS 100 (remote environmental monitoring Unit) AUV (Fig. 9) to map and define the boundaries of *Zostera marina* (eelgrass) close to the entrance of Sequim Bay, USA. Instead of raw sonar, return data image processing was performed on the GeoTiff images. These images were assigned with different histogram stretches at the pre-processing stage, and afterward the K-means unsupervised classification technique was applied to classify and detect eelgrass.

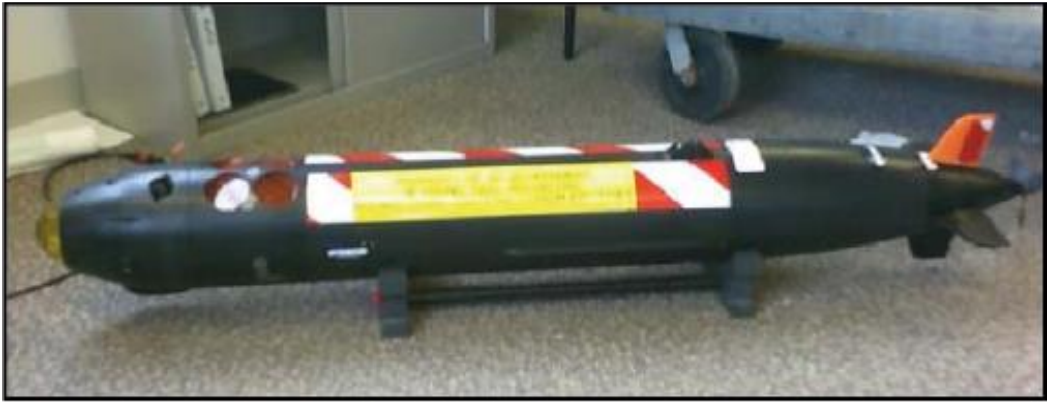

Fig. 9. REMUS AUV used in survey at Sequim Bay (Jones et al. [46]).

Lefebvre et al. [58] combined a profiling sonar with a sediment imager sonar (SIS) to detect and map *Zostera marina* in Calshot Spit, West Solent, on the south coast of England. The SIS was swept over the survey area and acquired acoustic images (.img file). Afterwards, these images were converted to an ASCII image (.xyz file) with the Sediment Imagery converter 1.0. For validation purposes, a video camera (Divecam-550c) was towed twenty metres behind the SIS and attached to a sledge from the survey vessel. Acoustic data processing and analysis were performed with Matlab®7.6, and the data was plotted with arcGIS®9.2[58]. An analysis of the acoustic data provided information on the presence of seagrass the canopy height and the species classification was usually performed by reviewing the video footage. A similar method was adopted by Paul et al. [81] on different sites to detect and map *Zostera marina*, *Zostera noltii*, and *Posidonia oceanica*.

Vasilijevic et al. [104] performed a study which involved the close monitoring, detection, and classification of seagrass (*Posidonia oceanica*) using an autonomous underwater vehicle (AUV). This approach was undertaken as part of the international interdisciplinary field training of marine robotics and application called 'breaking the surface' undertaken at Bay Lucia in the Croatian Coast of Murter. Vasilijevic et al. [104] used four AUVs for the field experiment and data collection. Out of these four AUVs, 2011-IVER2 (ocean server) (Fig. 10 (a)) is equipped with a Sportscan-Imagenex side-scan sonar and HERO2 underwater camera, 2012-IVER2 has a high-definition L-3 Klein's UUV-3500 side-scan sonar, 2012-LAUV (OceanScan) has a YellowFin-Imagenex side-scan sonar (Fig. 10 (b)), along with a digital camera, and 2013-REMUS100 (Hydroid) is equipped with EdgeTech 2205 high-definition side-scan sonar. Each of these AUV's used its own data visualisation software. The side-scan sonar mosaic of the study is shown in Figure 10 (c). The automatic bottom coverage estimation was calculated by an algorithm relying on underwater image brightness segmentation, developed by the University of Zagreb, and the seagrass identification was done by a human operator; also, the meadow density was calculated by the number of leaf shoots per square metre. This was also a fully manual labour-oriented process.

In a very recent work, Greene et al. [33] built and used a low cost side scan sonar array to map seagrass at lower depth areas (one metre or less) at Lower Laguna Madre in Texas, USA. The image quality from the sonar array was 2.5 cm/pixel. Validation of this approach was performed using satellite and aerial standard. Although it is a very cost effective way to collect acoustic data, it showed a number of drawbacks like, unreliability in noisy environment, high sensitivity to nearby vessel or towfish yaw, unavailability of commercial motion correction unit and over estimation of target size. Table 4 lists all of the discussed approaches which are based on underwater acoustic image data. Discussed articles do not provide any information on accuracy which is reflected on Table 4.

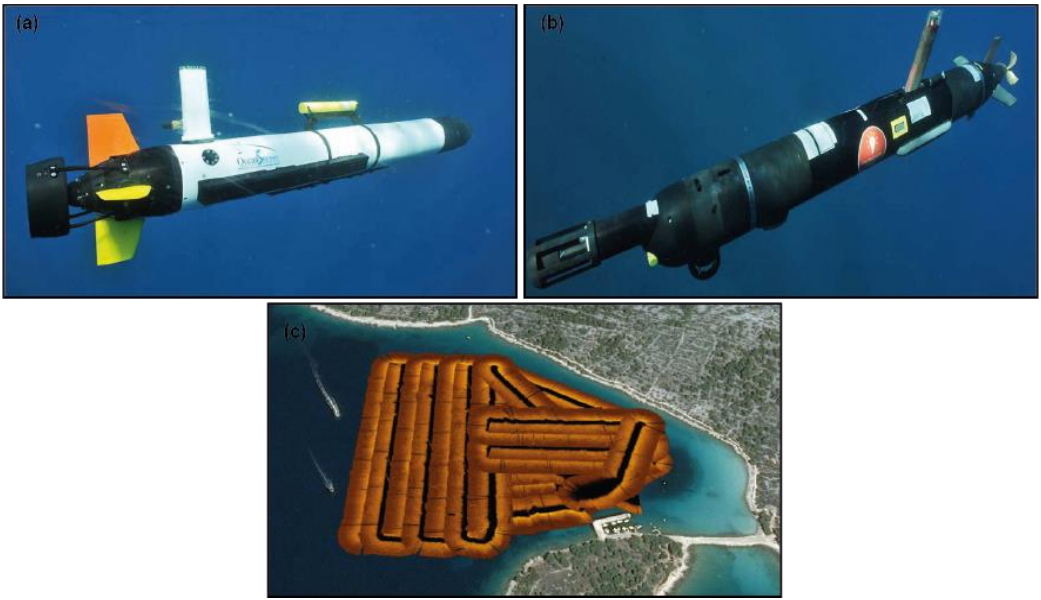

Fig. 10. (a) Klein-3500 side-scan sonar attached to AUV IVEr2, (b) Yellow-fin side-scan sonar attached to AUV LAUV, and (c) The study area and the AUV path (Vasilijevic et al.) [104]).

## 6 SPECTRAL IMAGE-BASED APPROACHES

Seagrass detection with spectral or hyperspectral imaging is one of the most popular methods in this research area since the data collection is faster than any other surveying techniques. Hyperspectral imaging creates a 3-dimensional data cube from spectral images of different wavelengths. This 3D cube is a combination of a 2D spectral image and information about the spatial pixel location. Spectral bands of this type of image dataset differ based on the used spectrometer, light source, and image sensor. For the interpretation of the spectral images, a spectral signature which is the variation of the reflectance of a substance with respect to the wavelength, plays a vital role. Knowledge of this signature helps to analyse an image, in some cases better than a conventional still image. Multimodality, noise and water penetration are the main limiting factors for spectral and hyperspectral image classification techniques.

In optical shallow water throughout the coastal areas, spectral images from remote sensors, such as sensors that are attached to modern satellites, sensors attached to an aerial platform, or even portable spectral sensors can be a very effective source of data that can be used to monitor, map, and characterise underwater vegetation. Moreover, recent image-processing techniques and computer-based classification algorithms have opened the door for new quantitative and machine-learning techniques which have revolutionised the old-fashioned photo-interpretation techniques in terms of time, cost, independence, and reliability (Wang [112]). Based on the imager (i.e. sensor) used for data collection, the seagrass detection and mapping approaches can be divided into four categories:

Table 4. List of seagrass detection approaches based on underwater acoustic image data

| Author | Data Type | Location | Source | Classifier | Accuracy |
| --- | --- | --- | --- | --- | --- |
| Vasilijevic et al. [104] | Acoustic Image | Croatian Coast of Murter Island | AUV with Sportscan-Imagenex side scan sonar | Manual | N/A |
| Lefebvre et al. [58] | Acoustic and Video | Calshot spit, West Solent on south coast of England | Sediment Imager Sonar (SIS) | Video Footage assessed by human expertise | N/A |
| Jones et al.[46] | Acoustic Image | Travis Spit, Sequim Bay | AUV with side scan acoustic imaging device | K-means unsupervised classification | N/A |
| Komatsu et al. [53] | Multi-beam Acoustic Image | Otsuchi Bay of Sanriku Coast, Japan | Multi-beam Sonar | N/A | N/A |
| Greene et al. [33] | Acoustic Image | Laguna Madre in Texas, USA | Side scan sonar array | N/A | 13.15%-22.18% overestimation |

approaches that use aerial platform based imagers (Section 6.1), approaches used satellite based imagers (Section 6.2), approaches based on portable imagers (Section 6.3), and approaches those worked with multiple imagers based images (Section 6.4). All these types are discussed below.

### 6.1 Aerial Imager Based Approaches

One of the initial approaches of mapping underwater vegetation was done by Haegele [34] between Ganges Harbour and French Creek, along the British Columbia coast. Colour infrared and colour 23 × 23 cm vertical aerial photos were taken. Haegele determined that underwater vegetation mapping is easier from aerial photographs where the tide is low and a colour-infrared film is used. From the infrared photographs, vegetation was classified from the colour patches that are acquired by the film and filter combination. Seagrass was identified and mapped using pinkish-red patches in the photo. Using a mirror stereoscope, flight lines were plotted from the photographs. The vegetation classification and mapping were done by colour pencil. The validation of the mapping and the classification of the right vegetation was done by the field survey, performed by scuba divers. The classifications were performed manually from infrared photographs.

Another early use of spectral images to detect seagrass was performed by Mumby et al. [77]. They used a compact airborne spectrographic imager (CASI) from an airborne platform at the coastal reef of the Turks and Caicos Islands in the British West Indies. The imager used eight spectral bands, and the spatial resolution was 1m per pixel. A calibrated visual scale was used to assess and identify standing seagrass crops. They discriminated their benthic habitat into two categories based on the field survey information: *coarse-level* habitats and *fine-level* habitat. The accuracy was respectively 89% and 81% for both types. This approach categorised seagrass under coarse-level habitat. This approach can only detect certain species of seagrass, corals, and algae, which limits its effectiveness.

To map and classify seagrass species (*Heterozostera/Zostera* and *Posidonia*) near the South Australian Bolivar Wastewater Treatment Plant, Anstee et al. [3] used hyperspectral images collected by CASI. After deducting the atmospheric effects, the images were fed to microBRIAN®software for classification. This software classified 99.7% of pixels into 256 basic classes and clustered them into 46 groups using the spectral tool 'Spectool'. Spectool calculated and compared the water reflectance at different depths on different substrates. Based on their expected reflectance signatures, they were able to map and classify *Posidonia* and *Heterozostera* with an accuracy rate of 72%. However, the presence of epiphytes affected the reflectance from the seagrass and complicated the classification which proved Fyfe and Dekker's [31] claim. Fyfe and Dekker [31] showed the true reflectance of seagrass can be affected and increased by the presence and growth of epiphyts. This approach also requires field survey and water-sample collection to obtain the optical property of the sample area so the reconstruction of the reflectance model in Spectool has higher accuracy. Table 5 lists the approaches that are based on aerial spectral images which were covered in this article.

Table 5. List of seagrass detection approaches based on aerial spectral imager data

| Author | Data Type | Location | Source | Classifier | Accuracy |
| --- | --- | --- | --- | --- | --- |
| Haegele [34] | Colour Infrared Image | Ganges Harbour and French Creek, British Columbia | Aerial Photography | Manual Classification | N/A |
| Mumby et al. [77] | Multi-Spectral Image | coastal reef of Turks and Caicos Islands in British West Indies | Compact Airborne Spectrographic Imager (CASI) | Calibrated visual scale | Coarse level 89% and Fine level habitat 81% |
| Anstee et al. [3] | Hyperspectral image | South Australian Bolivar Wastewater Treatment plant | Compact airborne spectrographic imager (CASI) | microBRIAN Software | 72% |

## 6.2 Satellite Imager based Approaches

*6.2.1 Landsat.* Although the standard was initially set by aerial photography to detect and monitor seagrass meadows, satellite spectral imaging techniques brought forth a cost effective solution for large area coverage. Landsat was one of the earliest to use map seagrass. Some of the inaugural events of using landsat satellite images for this purpose include approaches by Ackleson and Klemas [1], luczkovich et al. [64], Armstrong [4] and Ferguson et al. [25]. Landsat TM and ETM+ imagery was also used for coverage change estimation by Shapiro et al. [95].

In a later approach, Phinn et al. [91] combined remote sensing-based mapping and field survey-based mapping to increase the accuracy of classification at Moreton Bay. This time, the field data included spot-check photo transects with standardised labels and quadrate survey. A real-time video-camera feed (towed by the survey vessel) was interpreted and stored through Labview®software. This data was reinforced by a snorkel-based human observation. The seagrass species were confirmed through quadrant survey and photo transects collected by snorkelers. For remote sensing, a Landsat five thematic mapper multispectral image was used and classified using its reflectance signature by a minimum distance-to-means classification algorithm. To create a combined map, the authors used ESRI's ArcView software. This approach had a reliability of 83% for the coverage and 74% for the position. Though this approach has a high level of reliability, the spot survey is a complex and time-consuming approach which requires survey vessels, video interpretations, and photo transacts collection and interpretation.

For aquatic vegetation mapping at Honghu Lake in China, Li and Xiao [60] used a remote sensing technique based on spectral imagery. Their study site was Honghu wetland in Hubei province. For image data collection, they relied on the LAND ETM+ imager. Besides the depths and clarity of the water, the authors analysed the shape, texture, and spectrum of the seagrass distribution. Afterwards, they focused on the extraction of optimal features from images to enhance their classification accuracy. Li and Xiao [60] used their knowledge of mine classification to create a suitable classifier for seagrass. They built three classifiers, SVM, naive Bayesian, and decision tree. Li and Xiao [60] chose independent component analysis (ICA) to transform original images and used the independent components for classification. For classification, the authors performed a comparative analysis between decision tree, naive bayesian, and support vector machine (SVM) classifiers. Comparing these approaches, the overall accuracy of the Bayesian classifier was the highest (86.11%), and that of the SVM (85.90%) was the lowest.

*6.2.2 IKONOS.* The IKONOS 2 satellite was launched and carried the first commercial multispectral instrument which achieved a four metre spatial resolution [78]. One of the earliest approaches of using this high spatial resolution image was performed by Hochberg et al. [37]. The quasi-stochastic sea surface imposes serious accuracy concerns, where mapping underwater seagrass or other habitats using multispectral satellite images. To eliminate this effect and increase accuracy, Hochberg et al. [37] proposed an algorithm to remove glint from spectral satellite images. They applied their algorithm to a near-infrared band image of Lee Stocking Island (LSI), Bahamas, captured by the IKONOS satellite. They applied their glint-removal algorithm to the image and used a maximum likelihood classifier (MLC) with equal probabilities. Though the classification accuracy did not change after applying the glint-removal algorithm, the user's accuracy (true positive rate or the probability of a pixels assigned class is in-line with its actual class) significantly improved. For seagrass beds, the accuracy increased from 31.7% to 52.1%, and the boundary difference increased between seagrass and other habitats, such as coral, algae, and sand patches.

For the mapping of the benthic habitat of Roatan Island of Honduras, Mishra et al. [73] relied on high-resolution IKONOS multispectral data. To make their approach robust, they performed water-column and atmospheric correction and calculated water depth for each pixel, relying on a site-specific polynomial model. Their classification approach was based on the estimate of albedo (bottom reflectance). They found that an albedo ≥ 24% for sand patches, 12-24% for coral-dominated areas, and ≤ 12% for seagrass benthos. The bottom albedo was calculated using water optical properties and water depths obtained from IKONOS spectral data. They derived 150 different clusters from albedo using an iterative self-organising (ISODATA) algorithm. Finally, they classified the clusters into mixed coral, coral, mixed-seagrass, dense-seagrass, and deep water areas by an MLC. The accuracy of this approach was validated using digital video images from towfish and digital still photograph acquired by divers at 651 reference points. The overall accuracy of their method was 80.645%, but for dense-seagrass and mixed-seagrass environments, the accuracy was 75.82% and 72.52%, respectively. One of the major weakness is that this method solely depends on the light reflectance from seagrass and other benthic habitats. Therefore, the accuracy varies in the presence of microorganisms, coral bleaching, etc.

IKONOS multispectral images were also used for seagrass detection and mapping by Fornes et al. [27]. Their approach used high-resolution spectral images for *Posidonia oceanica* detection at the considerably clearer coastal area of Balearic Islands, Mallorca. The authors presented a method where a supervised classifier segregated *Posidonia oceanica*, rock, sand, and unclassifiable objects. This method was validated by an acoustical data-based survey. The overall classification accuracy based on the similarity with the acoustic survey result was 84%. The authors did not examine their approach in deep water or places with low water clarity. Moreover, ground truthing was not done for validation.

*6.2.3 QuickBird.* QuickBird was launched in 2001 and could collect both panchromatic and multispectral imagery before it was decayed in 2015. The resolution of multi spectral images (four bands) from quickbird was 2.44-1.63 meter [2]. Yang et al. [117] used a high-resolution multi-spectral Quickbird image to detect seagrass in the Xincun Bay, Hainan province, China. The bottom sun reflection had a significant impact on the accuracy level of seagrass detection, especially in shallow optically clear water. Yang et al. [117] used the radiance-transfer model to retrieve the bottom reflectivity. They used the relationship between the leaf area index (LAI) and the hyperspectral signal to process the Quickbird Image for mapping submerged seagrass. This experiment shows that spectral bands at 550 nm, 650 nm and 675 nm are significantly sensitive to LAI. ENVI and Photoshop were used during classification. For the determination of the classification-accuracy, the authors compared the Quickbird image pixels with situ ground truth of an area of 100$m^2$. The detection accuracy was more than 80% with Quickbird images. Their results also suggest that Quickbird images are easier for seagrass detection compared to Landsat images.

*6.2.4 WorldView-2.* WorldView-2 is the third and the most used satellite for seagrass mapping among the three satellites launched by DigitalGlobe. Four new spectral bands: Coastal, Yellow, Red Edge and Near Infrared 2 enabled this satellite imager to provide higher potential for vegetation species mapping. The resolution of the WV-2 satellite sensors are also higher (2meter for multispectral bands) [42]. For bathymetry and benthic habitat mapping at the Puerto Morelos Reef National Park of the Mexican Caribbean, Cerdeira-Estrada et al. [10] used this high spatial resolution WorldView-2 images and incorporated a physics-based data processing technique namely, EOMAP's modular inversion and processing system (MIP). MIP image processing was used on two WV2 satellite images of 2 m spatial resolution and eight multi-spectral bands for adjacency effect and sun-glitter correction. MIP also includes the water and atmospheric constituent's retrieval algorithms. Along with the spectral unmixing of the bottom reflectance, the sea floor albedo and water depth were calculated. For a robust habitat mapping, they used a two-level classification scheme relying on both geomorphological and biological characteristics. Through a spectral discrimination by segmentation process with eCognition software, the coverage of the habitats were detected. Subsequent habitat classification was based on the geomorphological characteristics. They classified the habitats into seagrass, microalgae, mixed vegetation, coral community, and sediment. For validation and ground-truth estimation, they conducted both optical and visual field surveys. Finally, for the accuracy calculation, a linear regression analysis was done on two subsets based on depths. This approach provides information on the presence of seagrass and other vegetation types, but does not classify the seagrass itself. After the segmentation of the satellite image, the whole classification technique relies on human expertise.

Eugenio et al. ([24]) used both multi-spectral (1.84 m) and panchromatic (0.46 m) images for their approach using WV2 at Canary Islands, North West African coast. Later these were re-sampled to 2.0 m and 0.5 m respectively. For their image dataset, the nominal swath width was 16.4 km. At the image-preprocessing stage, the authors (Eugenio et al. [24]) used a 6s atmospheric correction model which is formed by radiative transfer theory (Vermote et al. [105]; Svetlana et al. [54]). To eliminate the effects of the sea-surface from panchromatic and spectral high-resolution images, the authors combined both image processing and physical principles based techniques. After deglinting, a histogram matching was performed to equalize the images, statistically which is shown in Figure 11 (a). To eliminate the effects of large waves in the spectral images, the authors performed further improvements to their glint-removal algorithm. Figure 11 (b) shows the seabed reflectivity after surface reflectivity elimination. Afterwards, water column correction was performed. As multi-band spectral images were collected, in order to eliminate any misalignments in-between bands, the authors used a template matching (Vermote et al. [105]) technique. Finally, to classify the processed images, SVM was employed. The classification is shown in Figure 11 (c). The authors used Jeffries-Matusita metric (Canty [8]) for separability assessment.

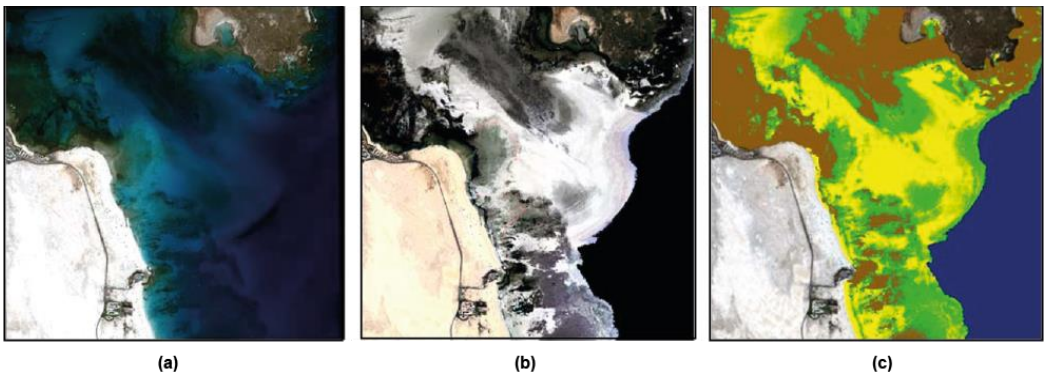

Fig. 11. Images of Corralejo area: (a) A colour image after deglinting and atmospheric correction, (b) Image after water coloumn correction (shows reflectivity of the seabed), (c) Final image for classification (blue = deep waters; yellow = sand; brown = rocks; green = seagrass) (Eugenio et al. [24]).

Seagrass and other benthic habitats around Fuerteventura Island (Corralejo) and Gran Canaria Island were mapped from multi-spectral ortho-ready images of the WorldView-2 satellite by Marcello et al. [67]. The image pre-processing stage of their approach performed water column, atmospheric, and sun-glint correction, and radiometric conversion. Marcello et al. [67] experimented with the maximum-likelihood classifier (MLC), SVM, mahalanobis distance (MH), and spectral angle mapper (SAM), and found the SVM with a Gaussian radial-basis kernel provided maximum accuracy. The kappa value for SVM was 0.71/0.79, which was the highest among other classifiers. The validation was done by in-situ data collection around Gran Canaria Island. In this approach, the authors thoroughly adjusted their parameters, such as the kernel type, the gamma terms, and the error penalty for SVM only, which made the higher kappa value for SVM questionable. Table 6 lists all the approaches covered in the article which relied on satellite spectral image data.

### 6.3 Portable Imager based Approaches

For the imaging of shallow water coastal areas, the ocean portable hyperspectral imager for low-light spectroscopy (Ocean PHILLS) is a specially designed spectral sensor by naval research laboratory (NRL). It is a lightweight compact instrument, designed from commercially available components that made it a cheaper option for land, water and sea floor mapping through spectral imaging [15]. Dierssen et al. [19] used Ocean PHILLS for the detection of seagrass and estimation of leaf-area index (LAI) *Thalassia testudinum* meadows close to Lee Stocking Island, Bahamas. To classify the benthic habitats, the authors applied a mechanistic radiative transfer approach, which removes the effect of water-column error and recovers the bottom reflectance for classification purposes.

Table 6. List of seagrass detection approaches based on satellite mounted spectral imager data

| Author | Data Type | Location | Source | Classifier | Accuracy |
|---|---|---|---|---|---|
| Phinn et al. [91] | Field Survey Still and Vedio Image with multi-spectral Satellite image | Moreton Bay, Australia | Towed Video Camera, Still Photography and Landsat 5 Thematic Mapper | Supervised Classifier and minimum distance to means Algorithm | Reliability 83% and 74% |
| Li and Xiao [60] | Spectral imagery | Honghu Lake, China | LANDSAT ETM+ image | Decision tree, Native Bays and Support Vector Machine (SVM) classifiers | Native Bays-86.11%; SVM-85.90% |
| Hochberg et al. [37] | Multispectral satellite image | Lee Stocking Island (LSI) | IKONOS Satellite Image | Maximum likelihood classifier with equal probabilities | 52.10% |
| Mishra et al. [73] | multispectral data | Roatan Island of Honduras | IKONOS spectral Imager | Maximum likelihood classifier | Overall 80.645% |
| Fornes et al. [27] | Multispectral Satellite Image | Balearic Islands, Mallorca | IKONOS Imager | Suppervised Classifier | 84% |
| Yang et al. [117] | High resolution multispectral Image | Xincun Bay, China | Quickbird Satellite Image | ENVI and Photoshop Software | N/A |
| Cerdeira-Estrada et al. [10] | Multispectral Images | Puerto Morelos Reef | WorldView-2 | eCognition Software | N/A |
| Eugenio et al. [24] | Multispectral Images | Canary Islands, North West African coast | WorldView-2 satellite | Supervised SVM | N/A |
| Marcello et al. [67] | Multispectral Ortho-ready images | Fuerteventura Island (Corralejo) and Gran Canaria Island | WorldView-2 satellite | SVM, SAM, MH, and MLC | N/A |

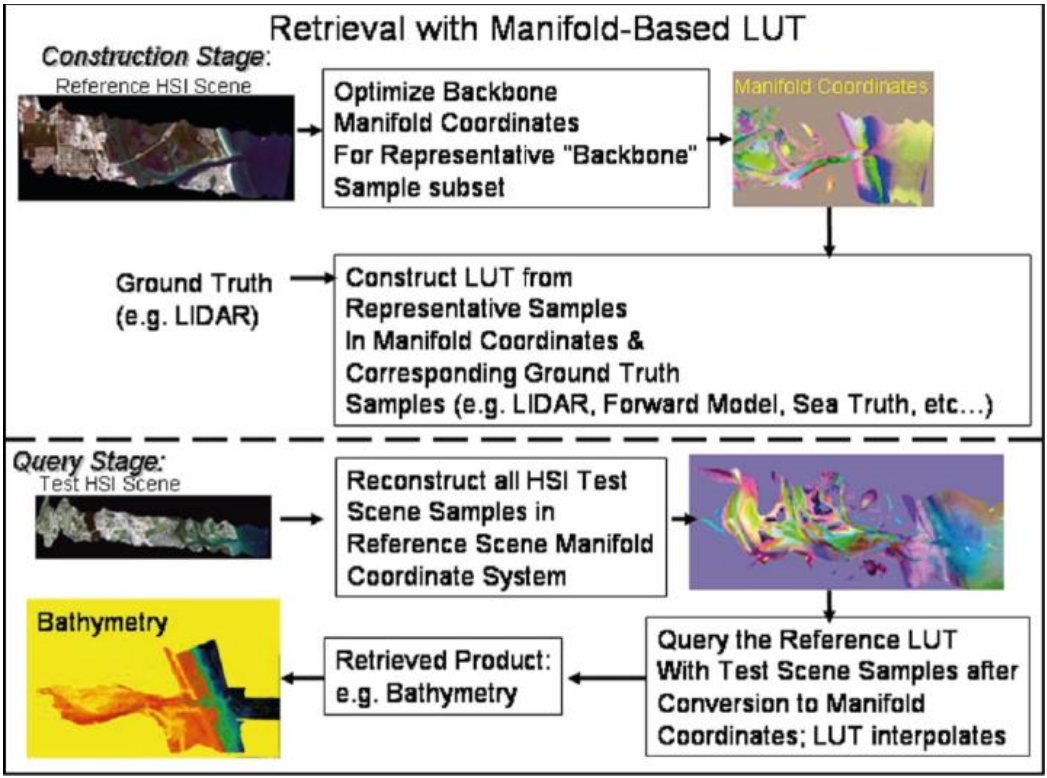

Fig. 12. The processing flow of Bachmann et al. [6]. (Top) LUT Construction with optimizing by manifold coordinates ground truthing by LIDAR data. (Bottom) Reconstruction of manifold coordinates and creation of query to the table that returns weighted depth or closest associated depth in case of multiple neighbours requested [6].

For their approach of seagrass detection and bathymetry mapping from hyperspectral images, Bachmann et al. [6] used a technique called manifold coordinate representations (MCR). For the data source, they used images from the Indian River Lagoon (IRL), located on the Florida's eastern seaboard, which is a 156-mile estuary. Hyperspectral images from 128 channels were collected using PHILLS camera. In this research approach, the accuracy of the manifold coordinate representations was examined as a reduced representation of a hyperspectral imagery (HSI) look-up table (LUT) for bathymetry retrieval (Fig. 12). The reason for choosing manifold coordinates is that, they are an intrinsic coordinates set and can parameterise naturally. Table 7 lists the portable imager based seagrass mapping approaches that are covered in the article.

### 6.4 Multiple Imager based Approaches

Seagrass detection and mapping with multiple spectral imagers is a very common practice by research in this area. One of the earliest approaches are by Mumby and Edwards [78]. They used IKONOS imagery for seagrass mapping and compared their results with other satellite imagers, namely, a thematic mapper, a multi-spectral scanner, and a CASI imager. They showed that the high-resolution IKONOS imagery can increase the thematic accuracy but fails to discriminate between different habitat classes.

Table 7. List of seagrass detection approaches based on portable spectral imager data

| Author | Data Type | Location | Source | Classifier | Accuracy |
| --- | --- | --- | --- | --- | --- |
| Dierssen et al. (2003)[19] | Hyperspectral Image | Lee Stocking island, Bahamas | Ocean PHILLS | N/A | N/A |
| Bachmann et al. (2009)[6] | Hyperspectral Image | Indian River Lagoon (IRL), Florida | Ocean PHILLS | Lookup table (LUT) | N/A |

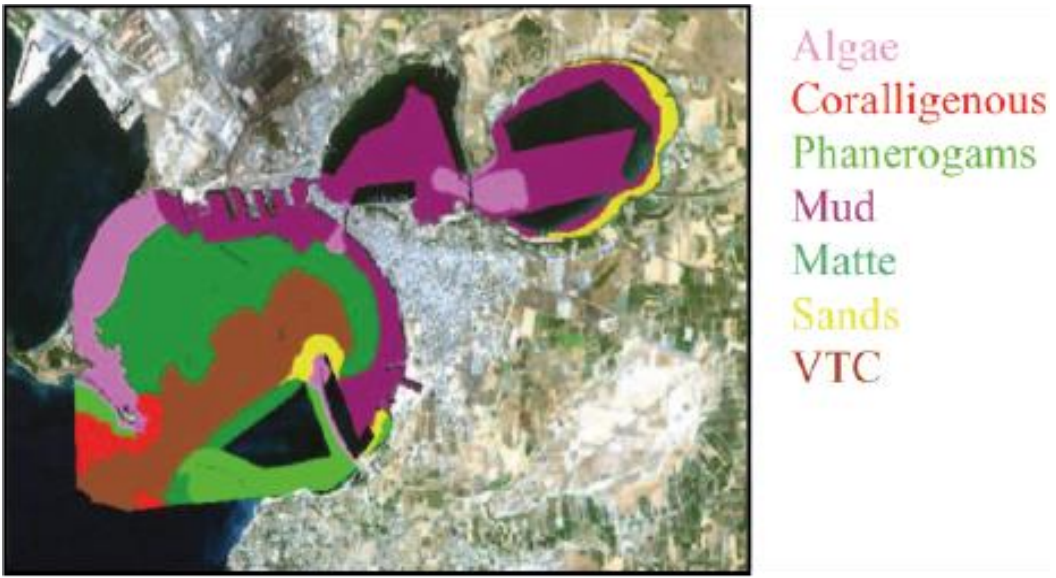

Fig. 13. Map of benthic habitats of Taranto Gulf (Matarrese et al. [69]).

For mapping *Posidonia oceanica* at the Taranto Gulf of Ionian Sea in Italy, Matarrese et al. [69] examined the potential of advanced spaceborne enhanced thematic mapper plus (ETM+), the advanced spaceborne thermal emission and reflection radiometer (ASTER), and the multispectral IKONOS imager. The authors did not perform any atmospheric correction for this approach. Rather, they converted the pixel values for all their data to radiances and applied a cloud and land mask. For classification purposes, the authors applied supervised maximum likelihood classifier in two stages: initially without bathymetric information and later with bathymetric information. Their experimental results show a significant accuracy improvement when classifying with bathymetric information. The accuracy for IKONOS, ASTER, and ETM+ with bathymetric information was 61%, 62%, and 70% respectively. Figure 13 shows their final map for the benthic habitats of Taranto Gulf.

Phinn et al. [85] assessed and compared the performance of hyperspectral aerial image-based detection and the multispectral satellite image-based seagrass detection while mapping seagrass species in Moreton Bay, Australia. For both image types, the authors used a minimum distance-to-means algorithm for classification through a supervised classification process. In this approach, Quickbird-2 and Landsat-5 thematic mapper images were used to represent the multispectral image, and a CASI-2 sensor provided hyper-spectral airborne images. Considering each mapped parameter, the hyperspectral images provided a higher accuracy of 46% compared to the multispectral images. This approach and their used classifier were unable to provide a higher accuracy level. Their assessment was heavily dependent on a manual survey and did not provide the attributes for seagrass classes.

Chen et al. [12] proposed to map the coastal and marine habitat of Malacca using Landsat and SPOT satellite images. They mainly used SPOT-5 satellite images during low tides, with a 10 m spatial resolution, when cloud-free data were unavailable from SPOT-5, Landsat thematic mapper and SPOT-4 images were collected. At the image pre-processing stage, the top of atmosphere (TOA) spectral radiance conversion was performed by the band calibration coefficients, and Rayleigh scattering and gaseous absorption correction were performed. On each Image, they applied an unsupervised hierarchical classification method. They used the ISODATA algorithm to differentiate between the dominant spectral clusters from the spectral reflectance. Finally, based on their visual interpretation, the clusters were classified into thematic categories. For the evaluation of their approach, they compared WorldView-2 and GeoEye data. In this approach, their map indicates the presence of seagrass or algae along with coral reefs, sand, and mangroves but does not clarify their exact class. For the lower densities of seagrass or algae, the SPOT and Landsat images even proved less effective. The accuracy of their SPOT and Landsat data is also ambiguous.

For detecting and mapping seagrass, as well as determining information regarding the seagrass-based biomass at Merambong Shoals in the Strait of Johore, Malaysia, Hashim et al. [35] used satellite imagery from the Landsat-8 operational land imager (band 2-blue, 3-green, and 4-Red). After geometric correction, sun-glint removal, atmospheric correction, and water column correction, the data was further processed to recover substrate-leaving radiance. Afterwards, the band reflectance index (BRI) was calculated using that radiance for all three bands. Finally, the MLC was appointed to map the existence of seagrass on the study area. The calculation of seagrass biomass was also performed using the satellite data and the biomass collected from the field survey. In this approach, they also attempted to show the relationships between seagrass coverage and seagrass-based biomass. The accuracy of their seagrass and non-seagrass classification is 90% overall. This method can only detect the existence of underwater vegetation, not its type. The authors assumed that the water attenuation coefficient was constant for all types of benthic habitats. In practice, the reflectance and attenuation vary according to the place, time, depth, and vegetation types. This approach also requires a field survey to calculate the biomass information.

Matta et al. [70] mapped underwater vegetation in the Gulf of Oristano, Sardinia, using multi-spectral images from KOMSAT-2, MIVIS, and RapidEye satellite images. The authors removed atmospheric effects using the 'Second Simulation of the Satellite Signal in the Solar Spectral' (6S) code and finally used a low-pass filter to reduce the residual noise and BOMBER for the mapping of *Posidonia oceanica*. Their approach achieved 88% accuracy.

To prove the effectiveness of using the spectral response through remote sensing from underwater vegetation for classification, Tin et al. [101] conducted an experiment to determine the spectral characteristics of some submerged aquatic vegetation which included seagrasses, macroalgae, and sand. In this study, they attempted to create a spectral reflectance profile library for 22 species of aquatic vegetatikon, including two different seagrasses and brown, red, and green macroalgae, as well as for sand and rubble. Using a high-resolution FieldSpec®4 Hi-Res portable spectro-radiometer at 350 nm to 2,151 nm in wavelength, the spectral reflectance was measured under clear skies. Statistical techniques such as correlation, spectral clustering, one-way ANOVA and principle component analysis (PCA) techniques were used on an IBM SPSS 20 platform to measure the reflectance difference. Though PCA showed significant differences of reflectance for green, brown, and red macroalgae groups, seagrasses like *Amphibolis antartica* and *Posidonia sp.* were misclassified and mixed up with red and green macroalgae (Fig. 11). This standard laboratory set-up used a hyperspectral measurement at 1 nm, which is not possible for the existing multi-spectral satellite sensors such as WorldView-2, IKONOS, Quick Bird, and Landsat. Moreover, in this approach was undertaken in the absence of a water column, while in practice, the water columns, water turbidity, and other environmental factors would significantly affect the reflectance profile.

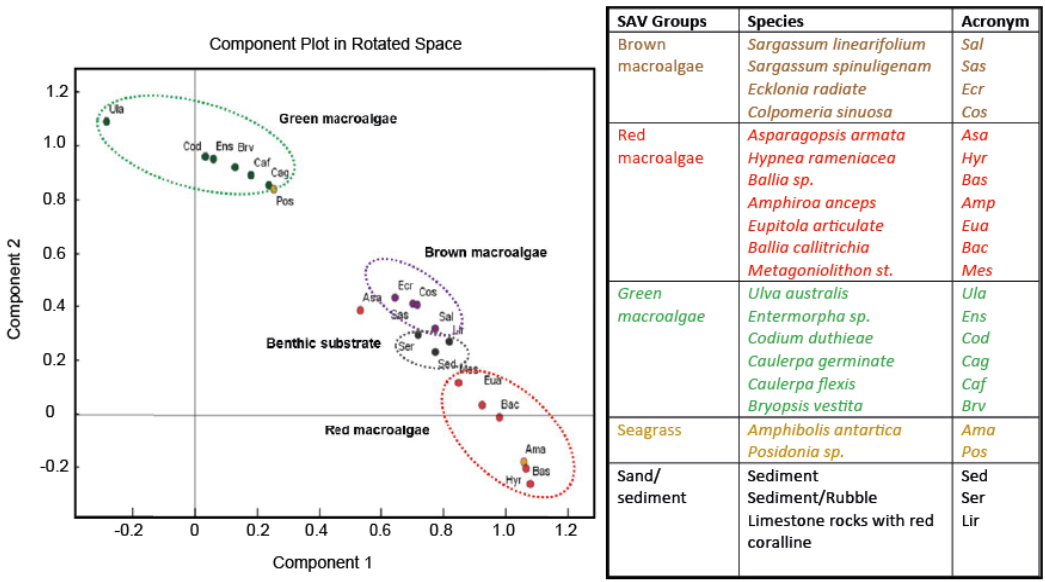

Fig. 14. After PCA, five vegetation types are segregated through a scatter plot: sand and sediment in 'black', seagrass in 'orange', green macro-algae in 'green', red macroalgae in 'red', brown macroalgae in 'violet'. (Tin et al. [101]) (best seen in colour).

To quantify the abundance of seagrass along the western coastline of Pinellas County, Florida, Pu and Bell [88] utilised images from Landsat-5 TM, hyperion (HYP), and Earth Observing-1 (EO-1) satellite sensors. In this approach, the Vegetation cover was classified into five different classes using a maximum-likelihood classifier. For biometric estimations, the leaf area index (LAI), percentage of SAV, and biomass were calculated. Finally, for the generation of an abundance map, the authors used a technique called fuzzy synthetic analysis. Pu and Bell [88] concluded that the HYP spectral sensor provided the best outcome during classification with an accuracy of 87%.

Using another approach, Pu and Bell [89] compared two different satellite images in terms of seagrass mapping and classification. They explored the competency of IKONOS satellite data at 4 m resolution from the IKO sensor for seagrass detection, mapping, and percentage cover calculation by comparing the results with Landsat TM images of the same study area, the mid-western coast of Florida, USA. Their analysis methodology consisted of three stages: image pre-processing for depth-invariant bands (DIBs) calculation, textural information extraction, and analysis of seagrass spatial distribution patterns. For DIBs calculation, all visible bands at surface radiance were calculated. Pu and Bell [89] used MLC and SVM for seagrass classification. Finally, a seagrass spatial distribution pattern was calculated. Using both classifiers, the study area was classified into <25%, 25-75%, and >75% vegetation coverage. For accuracy calculation, kappa (K), and overall accuracy (OA) were used as performance measures. In their approach, the seagrass detection accuracy was 5-20% higher from the IKO sensor data compared to previous works. Table 8 summarises all the multiple imager sourced image-based approaches described in this section.

Table 8. List of seagrass detection approaches based on multiple Spectral imager data

| Author | Data Type | Location | Source | Classifier | Accuracy |
| --- | --- | --- | --- | --- | --- |
| Mumby and Edward [78] | Multispectral, panchromatic & Thematic | Turks and Caicos islands | MSS, TM, SPOT, CASI & IKONOS | Supervised Classification | Heighest 81% |
| Matarrese et al. [69] | Multispectral Image | Taranto Gulf of Ionian sea in Italy | ASTER, ETM+ and IKONOS imager | Supervised maximum likelihood classifier | IKONOS-61%; ASTER-62% & ETM+-70% |
| Phinn et al. [85] | Multispectral and Hyperspectral Images | Moreton Bay, Australia | QuickBird-2, Landsat-5 and CASI-2 Airborne Imager | Minimum distances to means algorithm | 46% |
| Chen et al. [12] | Multispectral Images | Malacca | SPOT-5, Landsat Thematic Mapper and SPOT-4 | Visual interpretation of clusters | N/A |
| Hashim et al. [35] | Spectral imagery | Merambong Shoals in Strait of Johore, Malaysia | Landsat-8 Operational Land Imager | Maximum likelihood classifier | Overall 90% |
| Matta et al. [70] | Multispectral image | Gulf of Oristano, Sardinia | KOMSAT-2, MIVIS and RapidEye satellite images | BOMBER | 88% |
| Tin et al. [101] | Hyperspectral measurement at 1nm | Shoalwater Islands Marine Park at Rockingham, WA | FieldSpec®4 Hi-Res Portable spectroradiometer | Spectral clustering, One-way ANOVA and Principle Component Analysis (PCA) | N/A |
| Pu and Bell [89] | Multispectral Images | North-western coastline of Pinellas, Florida | IKONOS IKO sensor & TM of Landsat Satellite | Support Vector Machine (SVM) and Maximum Likelihood Classifier (MLC) | N/A |

## 7 CHALLENGES AND FUTURE RESEARCH

### 7.1 Challenges in Imaging technique Selection

Choosing a suitable and effective survey strategy, considering the goal and purpose of the survey itself is a challenge. Remote sensing is not always effective or even possible [41, 62]. When the objective is to detect and map dense seagrass meadows in cooler, clearer water, satellite and airborne sensor-based spectral imagery are very effective, but in warmer regions, where the water turbidity is higher, the effectiveness and accuracy are compromised [71]. When the depth of a coastal area is more than 2-3 metres, the 2-D digital spatial photographs taken from an aircraft become less credible for species-diversity mapping as blue light penetrates at a depth of more than 3 meters only [71]. For quick and large area mapping, underwater digital photography or video collection is considered too time-consuming [79].

### 7.2 Challenges in Underwater Image Based Seagrass Mapping

The most significant challenge of large area mapping with underwater photography is the relatively slow data collection procedure. Position stamping underwater images is also a major challenge. For the classification, most of the 2D underwater image-based approaches have used machine-learning classifiers such as SVM, SHIFT, LMT, RF, HMAX, and CQ-HMAX. Mostly, their task was to differentiate underwater vegetation from the surrounding environments, usually a mix of corals, sand beds, sponges, etc. Those approaches faced limitations to differentiate between seagrass, seaweed, and algae. Although, Gonzalez-Cid et al. [32] and Massot-Campos et al. [68] proposed to detect a specific seagrass type: *Posidonia oceanica*, no approaches detect and classify seagrasses from a dataset of mixed seagrass classes. We suggest that, deep machine learning techniques may be used for an automatic seagrass detection, segmentation, and classification from 2D or 3D underwater images, but deep convolutional neural networks require thousands of labelled image data to train a suitable deep network architecture.

### 7.3 Challenges in Video Data Based Seagrass Mapping

For deeper area mapping where the still or spectral images are not fully effective, video footage-based seagrass survey approaches provide some positive aspects, such as high confidence intervals, better sampling efficiency, positive identification of different seagrass species, and a dataset not only for seagrass mapping but also for other benthic habitat monitoring. But these video data based techniques have some serious drawbacks, such as digital video data needing a higher computation facility in general. Moreover, surveying a large area quickly with an underwater video camera is impractical. Norris et al. [79], and Lirman and Deangelo [61] used video files but the classification was performed by human experts. Few approaches have used video data just to validate their model based on other data formats not to derive their estimates of seagrass existance and coverage. So a bigger opportunity exists to detect, classify, and calculate seagrass density in real-time using video images collected by either AUVs or diver-based platforms (using a GPU-based computation system) and ANN algorithms. Also, the speed of the survey vessel and water quality are big limiting factors [96]. Further, video interpretation is a biasing factor when human expertise is the sole method for classification [14]. Therefore, there is an opportunity to apply automated image recognition routines to the video data.

### 7.4 Challenges in Acoustic Data Based Seagrass Mapping

The cost of sonar-based acoustic data-acquisition systems appears to have limited their application for underwater vegetation mapping. Though vegetation such as eelgrass can easily be detected with sonar images, it is very difficult to differentiate it from other seagrasses and other vegetation types such as macroalgae using acoustic data. Although the other distribution and coverage variance can be mapped with a side-scan sonar, this process is expensive, time consuming, the positioning is difficult, and the approach has to be adapted on a site-by-site basis. Furthermore, accuracy is drastically affected when the deapth range increase or the height of the vegetation increases [51]. A multi-beam sonar provides a 3D image of the seabed but is not suitable for shallow (<5m) regions [53], so its usability decreases for seagrass mapping.

### 7.5 Challenges in Spectral Data Based Seagrass Mapping

Spectral images are mostly acquired by satellite sensors. The cost of data collection is minimal, and large area mapping is possible within a very short time and using a limited number of images. Therefore, the largest number of seagrass detection and mapping approaches use multispectral images. However, spectral images from satellite or aerial sensors can only be used to detect the presence of underwater vegetation. For class (e.g. species) segregation or condition monitoring in situ data formats are required. Some of the latest satellites can provide relatively high-resolution image data. But in that case, the spatial area under single image decreases drastically. The validation of the existence of seagrass from satellite images still requires field survey, global positioning, and complex geo-fencing techniques. The classification techniques of satellite image data are mostly supervised machine-learning classifiers, including SVM, MLC, PCA which are still-semi automatic and supervised and ipractical to implement in real time detection and classification. Multispectral image datasets from satellites are also affected by the depth, along with the weather conditions, during data collection. Besides, The scale of mapping using both aerial and satellite sensor-based multispectral images are suitable for detecting large-scale changes but are less useful for accurate estimates of density (% cover), which requires high spatial resolution and hyperspectral data [85]. Seagrass mapping using aerial or satellite images requires ground-truth validation, which can be difficult for deeper medows [46].

### 7.6 Future Research Directions

Future research work for underwater vegetation mapping may shift more towards automatic data collection, detection, classification as well as validation which will replace the human dependent methods. While data collection is the main concern, there is a significant scope to work on cheaper and easier underwater image or video data collection techniques. Conventional machine learning approaches are based on semi-automatic feature extraction techniques. There is scope to use deep-learning based classification techniques which can potentially differentiate between seagrass patches from the surrounding environment and can distinguish individual classes in mixed seagrass frames as similar techniques have already been used for coral detection, classification and automated annotation in recent time [22, 66] .As the concept of 3D seabed mapping has already been considered [90], Creating 2D or 3D seabed maps with correct positioning stamps can also be a future research focus for both marine and the computer science research community. There are opportunities for research work on real time automatic percentage-calculation techniques as well.

## 8 CONCLUSION

Over the past two decades, significant improvements have taken place to develop the optimal technique of detecting and mapping underwater vegetation, especially seagrass. The attempts ranges from spatial monitoring to close monitoring. This paper surveys all the existing approaches and categorises them based on the type of images used. Discussions were made around two-dimensional spatial image-based approaches, two and three-dimensional underwater image-based approaches, video data-based approaches, acoustic image-based approaches, and spectral image-based approaches. This article also critically discussed the gaps of the most recent approaches which indicate that there are further scopes of research. As in recent years, deep learning has emerged as a new tool to solve computer vision problems and created a new hype among the research community; for seagrass detection, classification, and mapping, this promosing tool can be utilised for all data types. Three-dimensional imaging can be coupled with appropriate deep-learning architectures to create a new cutting-edge state-of-the-art standard for seagrass detection and seabed map generation.